\documentclass{ws-jmrr}

\usepackage[sort,compress,super]{cite}
\usepackage{cite}
\usepackage{url}
\usepackage{breakurl} 
\usepackage{wasysym}
\usepackage[usenames, dvipsnames]{color}
\usepackage{soul}
\usepackage{color}

\begin{document}

\markboth{Pre-edited version}{Accepted for publication in Journal of Medical Robotics Research on October 15, 2017.}

\title{Towards a Magnetically Actuated Laser Scanner for Endoscopic Microsurgeries}

\author{	Alperen Acemoglu\footnote{Corresponding author: Alperen Acemoglu,
					      	(1) 
					      	Department of Advanced Robotics, 
					      	Istituto Italiano di Tecnologia, 
					      	Via Morego, 30,
					      	Genoa, 16163, 
					      	Italy,
					      	{\tt\small alperen.acemoglu@iit.it.}
					      	(2) 
					      	Department of Informatics, Bioengineering, Robotics and Systems Engineering, 
					      	Universit\`a degli studi di Genova, 
					      	Via All'Opera Pia, 13, 
					      	Genoa, 16145,
					      	Italy.
					      	}, 
			Nikhil Deshpande and
			Leonardo S. Mattos}

\address{Department of Advanced Robotics, 
		Istituto Italiano di Tecnologia, 
		Genoa, 
		Italy\\
	}

\maketitle

\begin{abstract}
This article presents the design and assembly of a novel magnetically actuated endoscopic laser scanner device. 
The device is designed to perform 2D position control and high speed scanning of a fiber-based laser for operation in narrow workspaces. 
The device includes laser focusing optics to allow non-contact incisions and tablet-based control interface for intuitive teleoperation.
The performance of the proof-of-concept device is analysed through controllability and the usability studies.
The computer-controlled high-speed scanning demonstrates repeatable results with 21 $\mu$m precision and a stable response up to 48 Hz. 
Teleoperation user trials, were performed for trajectory-following tasks with 12 subjects, show an accuracy of 39 $\mu$m. 
The innovative design of the device can be applied to both surgical and diagnostic (imaging) applications in endoscopic systems. 
\end{abstract}


\begin{multicols}{2}

\section{Introduction}
The introduction of surgical capability as well as computer-assisted features to diagnostic-only instruments, 
allowed endoscopy to be incorporated into surgical thinking~\cite{litynski1999endoscopic}. 
Endoscopic surgical platforms are being utilized in various procedures including abdominal~\cite{litynski1999endoscopic}, transvaginal,~\cite{moris2012surgery} and trans-urethral~\cite{hendrick2014multi} due to their advantages or few-to-no incisions and closer access to surgical site.
Examples of these platforms include both manual and robotic systems: EndoSamurai from Olympus~\cite{endoSamurai}, DDES from Boston Scientific~\cite{ddes}, MASTER system~\cite{phee2009master}, the ViaCath system from Hansen Medical~\cite{abbott2007design}, and Anubis scope from Karl Storz~\cite{dhumane2011minimally}. 

Independently, the diagnostic capabilities of endoscopes have also been under investigation for augmentation. 
Researchers introduced optical fiber scanning traditional endoscope for capturing images~\cite{yelin2006three,fisher2009cobra}. 
A three-dimensional real-time imaging system using a single optical fiber which is scanned over the target area using an external mechanical-transduction mechanism is presented in~\cite{yelin2006three}. 
A similar fiber positioning concept is used in the Cobra fiber positioner device~\cite{fisher2009cobra}. 
Lee et al. discuss the scanning fiber endoscope which uses laser-based video imaging~\cite{lee2010scanning}. 
The design is based on custom piezo-electric actuators which position a cantilevered laser-light carrying fiber over the 2D imaging plane. 
Conceivably, this novel scanning fiber capability adapted to surgical lasers can provide a completely new approach to laser-based robotic endoscopic surgery~\cite{mattos2016microsurgery,renevier2017endoscopic}.

In the field of laryngology, the state-of-the-art non-invasive treatment technique is a laser-based procedure called Transoral Laser Microsurgery (TLM). 
TLM is a suite of complex laryngological methods involving the precise micromanipulation of a surgical laser in a narrow workspace (larynx and vocal cords). 
The goal is to perform tumor resection trans-orally, thereby avoiding the problems of traditional open-neck surgeries performed with cold steel instruments~\cite{rubinstein2011transoral}. 
Precise tissue ablations are needed to completely remove the malignant tissue while preserving healthy tissue around it.
In laryngology, free-beam lasers first became prevalent in the 1970's when the CO$_2$ laser was coupled to a surgical microscope~\cite{strong1972laser,strong1975laser}. 
The surgical setup includes a laryngoscope, a device that establishes and maintains a direct line-of-sight to the larynx under a surgical microscope (Fig.~\ref{fig_tradit}). 
The surgical laser is controlled with a micromanipulator, which is a small joystick, attached to the microscope. 
This micromanipulator controls the orientation of a beam deflection mirror which in-turn positions the laser spot at the  
\begin{figurehere}
	\begin{center}
		\centerline{\includegraphics[width=3.4in,trim={0 0 0 0}]{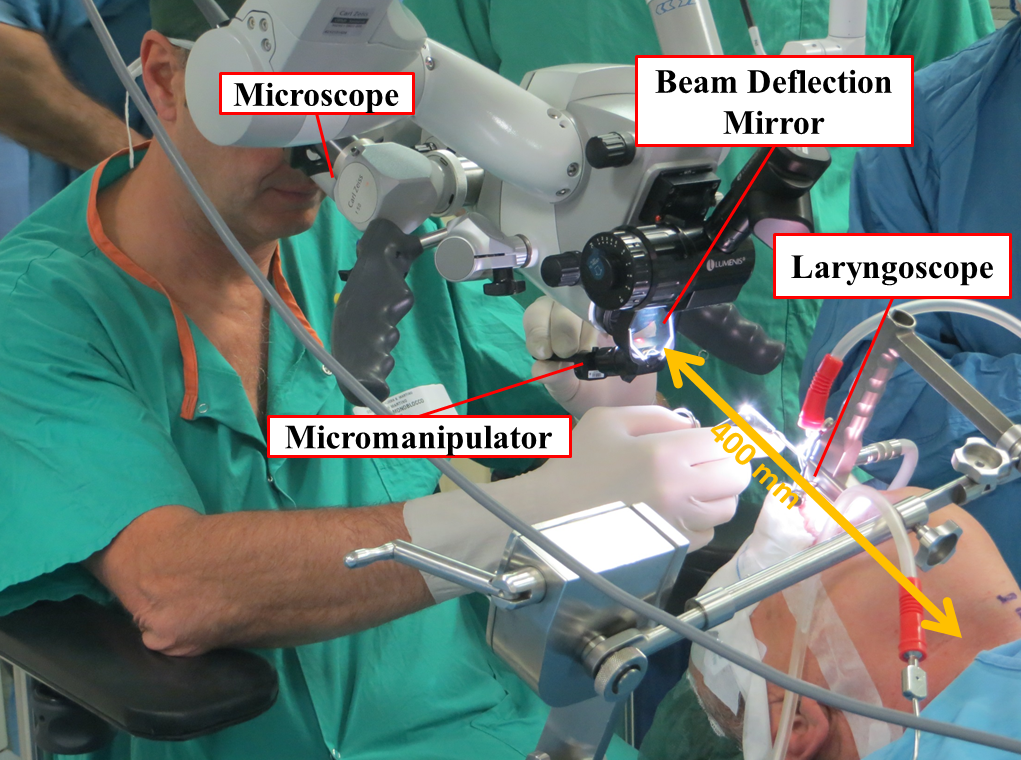}}
		\caption{The state-of-the-art of Transoral Laser Microsurgery (TLM) setup.}
			\label{fig_tradit}
	\end{center}
\end{figurehere}
surgical site.

This surgical setup is effective but suffers from significant limitations in controllability and ergonomics~\cite{mattos2014novel}. 
In these systems, the typical operating distance is 400 mm. 
This means that the microscope and the laser control device are located quite far from the surgical site, making the direct line-of-sight requirement during surgery highly challenging. 
The laryngoscope itself is a cumbersome device requiring the patient’s head to be put in an awkward position, which can cause soft tissue lesion, bleeding, or post-operative pains~\cite{kleine2016evaluation}.
Furthermore, proper usage of the micromanipulator is also highly dependent on the dexterity and experience of the surgeon~\cite{mattos2014novel, deshpande2013imaging, patel2012endoscopic}. 
Given these limitations, it is crucial to develop devices that can not only improve the delivery of the laser to the surgical site but also simplify the surgeon-machine interface while overcoming the challenges of controllability and ergonomics. 
The introduction of endoscopic surgery in TLM has the potential to address these limitations in a similar way as it has done in other procedures~\cite{deshpande2014enhanced,renevier2017endoscopic}.

\subsection{Endoscopic systems in TLM}
Performing laser incisions closer to the surgical site has high importance due to the advantages they can offer over the state-of-the-art systems. 
Endoscopic systems can eliminate the direct line-of-sight requirement and enable access to parts of the surgical site that are not reachable with the traditional systems. 
Reducing the operating distance can also potentially improve the precision of the surgical operations~\cite{patel2012endoscopic}: Laser position control can be performed close to the target tissue rather than from 400 mm away. 
Additionally, with a design adapted to the patient’s anatomy, endoscopic systems can eliminate requirement of the laryngoscope itself. 
On the other hand, the surgical microscope can be replaced with endoscopic cameras. 
These can provide simpler and more ergonomic visualization to the surgeons, and also introduce imaging-assisted features~\cite{renevier2017endoscopic}.
Lastly, the manual micromanipulator can be replaced with more intuitive control methods, e.g., a tablet device, providing a comfortable surgeon-machine teleoperation interface. 
The introduction of endoscopic surgical systems in laryngology has held interest in the research as well as commercial community. 
The FLEX Robotic System~\cite{remacle2015transoral} is the first flexible and steerable endoscopic system cleared by FDA for use in transoral surgical procedures. 
It offers a diameter of 19 mm and includes two manually-controlled, interchangeable, instrument arms as well as a camera for visualization. 
Researchers at Vanderbilt University have demonstrated two independent laparo-endoscopic systems: the IREP system~\cite{bajo2012integration} and the Concentric Tube-based system~\cite{hendrick2014multi}. 
The IREP offers integrated 3D vision within a 15 mm tube with two dexterous robotic arms with 6-DOF manipulation. 
The Concentric Tube-based robotic endoscopic system offers two continuum robot automated arms attached to a commercial endoscope for a small (dia. 8.66 mm) semi-automated system. 
Although none of these systems are designed for laser-based microsurgeries, their instrument arms can be adapted to carry optical fiber to provide laser incisions in contact with tissue.
Patel et al.~\cite{patel2012endoscopic} present another endoscopic tool designed for laryngeal resection using a CO$_2$ laser. 
This system delivers laser beam through an optical fiber to the tip of a 17 mm diameter customized endoscope and laser position control is performed by refracting the laser beam using two optical wedges. 
They presented the capability of the tool with different trajectories indicating that system can provide repeatable performance~\cite{patel2012endoscopic}.

Most recently, the European project $\mu$RALP has focused on developing an endoscopic laser scanning system for robot-assisted surgeries, enabling the replacement of the manual micromanipulator with a graphics tablet~\cite{mattos2016microsurgery}. 
Within $\mu$RALP, researchers presented an endoscopic tool consisting of a silicon-mirror based micromanipulation robot driven by two piezoelectric motors to steer a laser beam emanating from an optical fiber carried to the tip of the endoscope.
This endoscopic tool has 18 mm diameter and includes a stereo vision system. 
They also presented the image-based control of the laser spot position by placing optical fiber bundles to capture the images~\cite{renevier2017endoscopic}. 
As yet, the research has not incorporated any laser focusing optics in the tool for non-contact laser incisions. 

From this discussion, it is evident that a tool offering precise laser incisions in a small form-factor forms a critical component of any laser-based endoscopic surgical system. 
The research in this article takes its motivation from these previous research efforts to present a new endoscopic laser scanner device for robot-assisted endoscopic laser microsurgeries.

\subsection{Problem formulation}
The introduction of an endoscopic surgical device in TLM would require not only addressing the limitations of state-of-the-art systems, but also incorporating the features that surgeons have come to expect from TLM surgical devices. 
Therefore, it is essential to examine in detail the features of the state-of-the-art TLM systems.
As mentioned earlier, the current laser systems (e.g., Lumenis - Digital Acublade~\cite{acublade2017}, Deka HiScan Surgical~\cite{dekalaser}) use a small manual micromanipulator, attached to a surgical microscope, to steer a free-beam laser spot on the target tissue. 
By design, this implies that the incisions are non-contact.
The incorporation of a scanning system brings significant advantages with respect to quality of cut as well as thermal damage to and carbonization of surrounding tissue~\cite{remacle2005new}. 
This feature guarantees high incision quality with minimum carbonization and thermal damage during soft tissue ablations. Current commercial systems have high-speed scanning as a standard feature~\cite{acublade2017,dekalaser}. 
In free-beam systems, the scanning of the laser is performed with fast steering mirrors.

Consequently, the design of the endoscopic tool should involve: 
(i) delivery of laser to the surgical site while allowing non-contact incisions; 
(ii) incorporating the feature of high-speed scanning for high incision quality; 
(iii) improving the laser control interface; and (iv) providing  
additional features for improved surgical performance.
As has been discussed earlier, optical fibers can carry laser light to the tips of endoscopes~\cite{lee2010scanning}. 
Optical fibers can be coupled with high-power CO$_2$ and diode lasers to perform incisions in soft tissues~\cite{arroyo2016diode, fried2005high}. 
This ability allows the lasers to access the surgical sites that are difficult to reach with conventional methods. 
In addition, fiber-based lasers may contribute to shorten the operating time of some procedures as compared to free-beam laser surgery, as demonstrated for the case of stapes surgery~\cite{chang2017comparison}.

A limitation of fiber-based operations is that the incisions need to be performed in close proximity to the tissue, almost in contact~\cite{hendrick2014multi,remacle2015transoral}. 
This is because the laser light diverges as it exits the optical fiber. 
Additionally, existing fiber-based systems employ manual control of the fiber~\cite{patel2012endoscopic}. Due to the hand tremors, optical fibers often penetrate the tissue. 
The sticking of tissue residue to the fiber itself is a major problem, causing damage to the tip of the fiber and leading to reduced quality of the incision.
It would therefore be beneficial to isolate the bare optical fiber from the surgical site to avoid the direct interaction with the tissue and thereby retain the benefits of non-contact laser incisions.  
To overcome the problem of laser light divergence, optical lenses can be used at the distal end of the fiber to focus the light on a target tissue.
In order to enable the scanning feature in endoscopic applications, it is desirable to miniaturize the concept of the fast steering mirrors and adapt it to scanning the laser fiber itself.
Investigating new actuation mechanisms for scanning and position control is crucial for the surgical operations in the narrow workspaces.
Researchers have proposed magnetic actuation for the miniaturization of the surgical catheters~\cite{faddis2003magnetic,tunay2004position,boskma2016closed}. 
In these studies, a permanent magnet attached to a cantilevered beam is actuated with an external magnetic field for position control. 
Magnetic fields allow high frequencies appropriate for the scanning feature. 
Furthermore, the actuated cantilever beam can be replaced with a cantilevered optical fiber thus keeping the same control strategy with magnetic field for changing the laser beam direction.

Earlier studies have shown that the control of the laser with a manual micromanipulator is highly challenging especially during long surgeries~\cite{mattos2011virtual}. 
For a better control strategy, the tablet device control of the laser has been proposed providing a writing-like intuitive interface~\cite{mattos2016microsurgery,deshpande2013imaging,barresi2013comparative}. 
It has been shown that tablet device provides more intuitive and precise control for laser micromanipulation as compared to the traditional manual system.
Thus, miniaturization and adaptation of the state-of-the-art laser-based surgical systems can be done using the optical fibers and a tablet interface for endoscopic applications.

\subsection{Contributions of this paper}

This work aims to bring the benefits of free-beam lasers to laser-based endoscopic surgery by designing an end-effector module to be placed at the distal tip of an endoscope.
This is done through the design of a novel miniaturized magnetic laser scanner device.
The system provides 2D position control and high-speed scanning of a laser fiber using electromagnetic actuation. 
It focuses the laser on the target using optical lenses, thus
enabling non-contact incisions. 
The device is controlled using a graphics tablet with a stylus for intuitive laser tele-manipulation~\cite{mattos2012safe,deshpande2013imaging,barresi2013comparative}.
The features of this proof-of-concept device are analysed through characterization and teleoperation user trials with the device. 
Characterization experiments consist of validation of the high-speed scanning and repeatability of the position control mechanism. 
For user trials, the controllability and usability assessments of the magnetic laser scanner  performed in two parts: (i) objective analysis with the Root-mean-square-error (RMSE), maximum error, and execution time metrics; and (ii) subjective analysis with a dedicated questionnaire.

The article is constructed as follows: 
\textit{Section} II describes the design of the magnetic laser scanner. 
\textit{Section} III explains the characterization experiments and teleoperation user trials.   
\textit{Section} IV gives the details of the assessments methods.
\textit{Section} V presents the results of the experiments. 
\textit{Section} VI discusses the results.
Finally, \textit{Section} VII concludes the paper.

\pagebreak

\section{Design of the Magnetic Laser Scanner}
\subsection{Actuation mechanism}
The principle of the magnetic actuation mechanism is based on the interaction between magnetic field created by coils and a permanent magnet. 
When current flows through the coils, a magnetic field is induced in the region around coil.
For a single current loop, the magnetic field strength is given by Biot-Savart's law~\cite{smythe1988static}:
\begin{equation}
	\label{eq_mag_Bz_single}
	\mathbf{B} = \frac{ \mu_{0} I }{ 2 } \frac{ a^2 }{ ( a^2 + z^2 ) ^{ \frac{3}{2} } },
\end{equation}
\noindent where $\mu_{0}$ is the permeability of the free space, $a$ is the radius of the current loop,
$I$ is the current and z is the distance between the calculation point and the loop.
The induced magnetic field creates a force and a torque on any permanent magnet introduced in the field.
The magnetic force, $\mathbf{F_{m}}$, acting on a magnetic dipole moment, 
$\mathbf{m}$, can be expressed as the gradient of the magnetic 
potential energy, $U$.
The magnetic torque, $\mathbf{T_m}$, is the cross product of the magnetic dipole moment, $\mathbf{m}$, and the magnetic field:
\begin{subequations}
	\begin{align}
		\label{mag_force}
			\mathbf{F_{m}} &= \nabla{(\mathbf{m} \cdot \mathbf{B})},\\
		\label{mag_torque}
			\mathbf{T_m}   &= \mathbf{m} \times \mathbf{B},
	\end{align}
\end{subequations}
\noindent where $\nabla$ is the gradient operator.
The proposed magnetic laser scanner uses this principle to actuate an optical fiber within a specially designed endoscopic tool to perform 2D positioning and high-speed scanning~(Fig.~\ref{fig_endotool}). 
This tool is designed to be used as a tip section of an endoscopic system. 
The optical fiber is adapted to be a mechanical cantilever: the fiber is fixed at the center of the tool wall as seen in Fig.~\ref{fig_endotool}.
An axially magnetized ring permanent magnet is attached close to its fixed end.
In order to have the 2 DOF, two electromagnetic coil pairs are placed in x- and y-axes.
Since the laser light diverges when it exits the optical fiber, two plano-convex lenses are used to collimate and focus the laser on the target surface.
Thus, the magnetically actuated optical fiber changes the laser spot position on the target. 
The bending of the cantilevered optical fiber is primarily dependent on the magnetic torque. 
The effect of the magnetic force is neglected by assuming the magnetic field variations in the narrow workspace are small enough: 
since same current values are fed to the electromagnetic coil pairs, it is assumed that a uniform magnetic field is induced in the workspace. 
\begin{figurehere}
	\begin{center}
		\centerline{\includegraphics[width=3.5in,trim={0cm 0 0 0}]{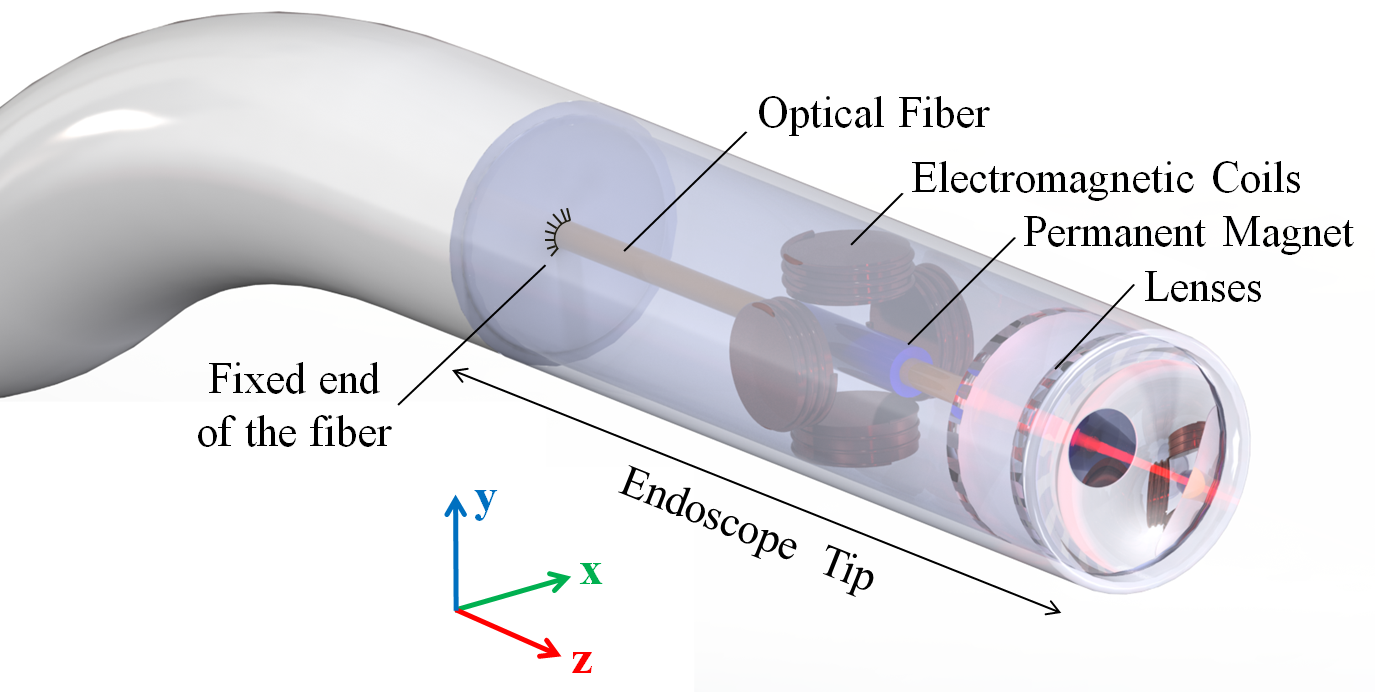}}
		\caption{Conceptual design of the Magnetic Laser Scanner. A system for the micromanipulation of an optical fiber based on magnetic interaction between electromagnetic coils and permanent magnet.}
			\label{fig_endotool}
	\end{center}
\end{figurehere}

\subsection{Prototype Manufacturing}
The mechanical design of the device is based on optimization of different constraints.
The main constraint is the overall diameter of the endoscopic tool.
The second constraint is the requirement of maximum magnetic field strength  to bend the cantilevered optical fiber so as to increase the total workspace of the tool: maximum number of turns must be wound on the electromagnetic coils to create maximum torque on the permanent magnet. 
In addition to these, four electromagnetic coils are placed for 2D position control of the laser spot. 
Thus, the optimization of the constraints can be defined as: \textit{placing four electromagnetic coils with maximum possible sizes inside an endoscope with minimum outer diameter}. 
In the current proof-of-concept design, the overall diameter of the tool is restricted to be less than 15 mm~\cite{renevier2017endoscopic, patel2012endoscopic}.
The magnetic laser scanner prototype presented here has 13 mm outer diameter with 5 mm core-diameter electromagnetic coils. 
The total scanning range achieved with this design is 4$\times$4 mm$^2$.
\vspace*{\fill}
\begin{tablehere}
	\tbl{Prototype Components of Magnetic Laser Scanner\label{tab_comp}}
	{\begin{tabular}{lc}
		\toprule
		Component Name           & Specification                                          \\ \colrule
		Laser source             & Visible red, $\lambda$ = 625nm 			              \\
		Multimode optical fiber  & $\diameter_{core}$ = 300$\mu$m,  NA = 0.39               \\
		Collimating lens         & $f$ = 12.5,  $\diameter$ = 9mm                   \\
		Focusing lens            & $f$ = 30,    $\diameter$ = 6mm                    \\
		Ring permanent magnet    & $\diameter_i$ = 0.7, $\diameter_o$ = 2, $h$ = 2mm \\
		Electromagnetic coils    & $N$ = 150 turns, iron core                             \\
		\botrule
	\end{tabular}}
\end{tablehere}
\begin{figurehere}
	\begin{center}
		\centerline{\includegraphics[width=3.5in,trim={0 0 0 0}]{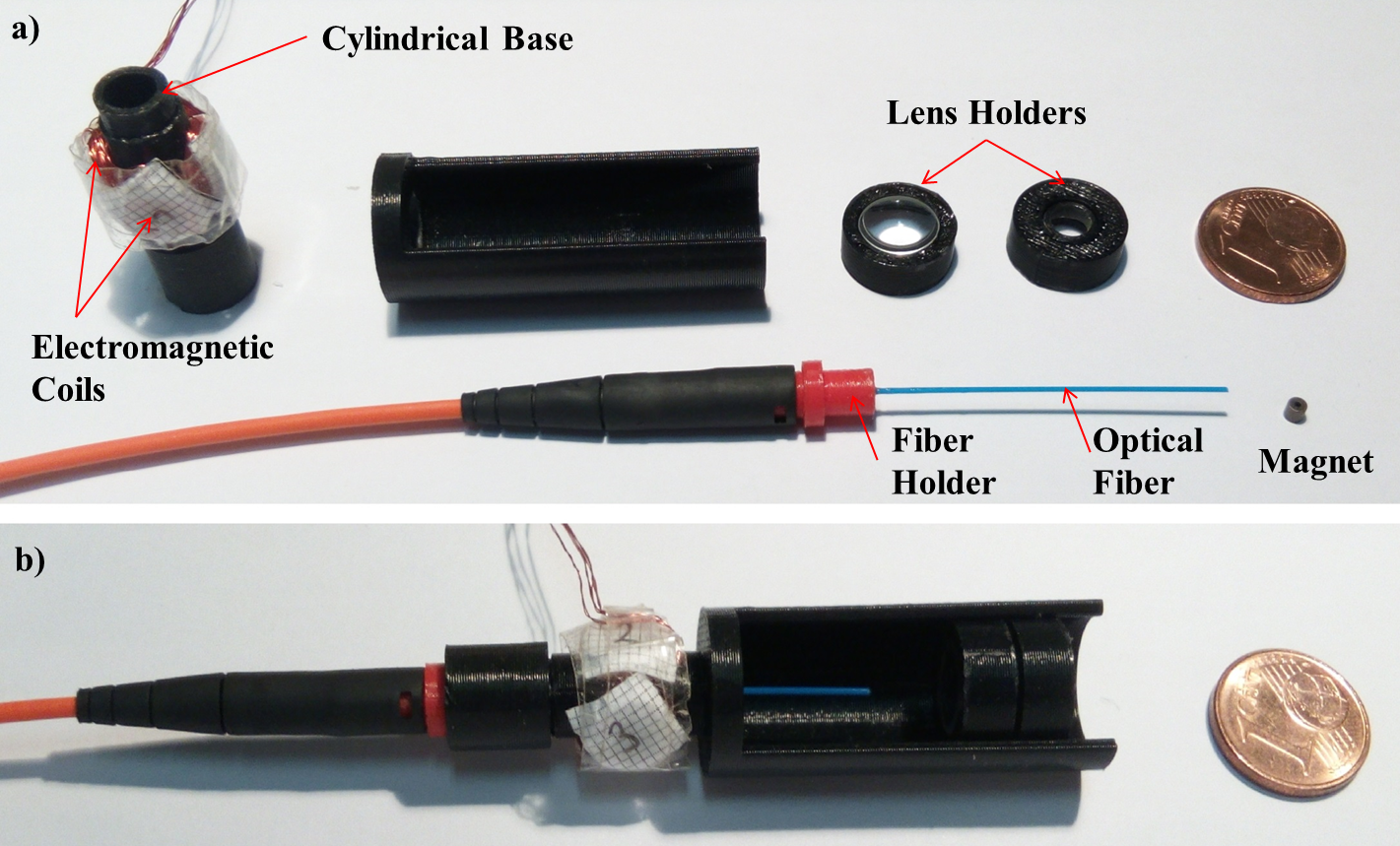}}
		\caption{Prototype components (a) and assembled prototype (b) of magnetic laser scanner.}
		\label{fig_manuf}
	\end{center}
\end{figurehere}
The proof-of-concept device was manufactured as follows:
The cylindrical base of the electromagnetic coils, lens and fiber holders were manufactured with 3D printing (see Fig.~\ref{fig_manuf}). 
Four ferromagnetic coil cores were placed circumferentially on this cylindrical base orthogonal to each other. 
Then, electromagnetic coils were wound on the cores with a thin copper wire.
The chosen copper wire allows a maximum current of $\pm$0.165 A.
Optical fiber was fixed to a fiber holder.
An axially magnetized ring permanent magnet was inserted over the optical fiber.
Lens holders were attached to the cylindrical base which holds the electromagnetic coils.
Table~\ref{tab_comp} presents component specifications of the magnetic laser scanner prototype.
The optical lenses were chosen to keep a 30 mm distance between the endoscopic tool and the target surface. 
This optical system produced a laser spot diameter of 0.57 mm on the target.\\

\subsection{Teleoperation control setup}
Fig.~\ref{fig_expset} shows the experimental setup for teleoperation user trials. 
The currents of the electromagnetic coil pairs placed in x- and y-axes are controlled with two analog outputs of an \textit{Arduino Due} control board, which is capable of receiving real-time commands at a rate of 4 KHz for controlling the analog outputs~\cite{arduino}.
The control board and a tablet device are connected to a computer via USB. 
These devices are interconnected by using Robot Operating System (ROS) to control the currents $I_x$ and $I_y$, with the tablet positions $p_x$ and $p_y$, respectively. 
This allows to map the tablet coordinates to the currents, $I_x$ and $I_y$:
\begin{equation}
	\label{eq_refraction_matrix}
	\begin{bmatrix} 
            I_x\\
	        I_y    
	\end{bmatrix} 
        =  
    \begin{bmatrix} 
            c_{11} & c_{12}\\
	        c_{21} & c_{22}    
	\end{bmatrix} 
	\begin{bmatrix} 
             p_x\\
	         p_y     
	\end{bmatrix} ,
\end{equation}
\noindent where $c_{11}$, $c_{12}$, $c_{21}$ and $c_{22}$ are four scalars.
\begin{figurehere}
	\begin{center}
		\centerline{\includegraphics[width=3.5in,trim={0cm 0 0 0}]{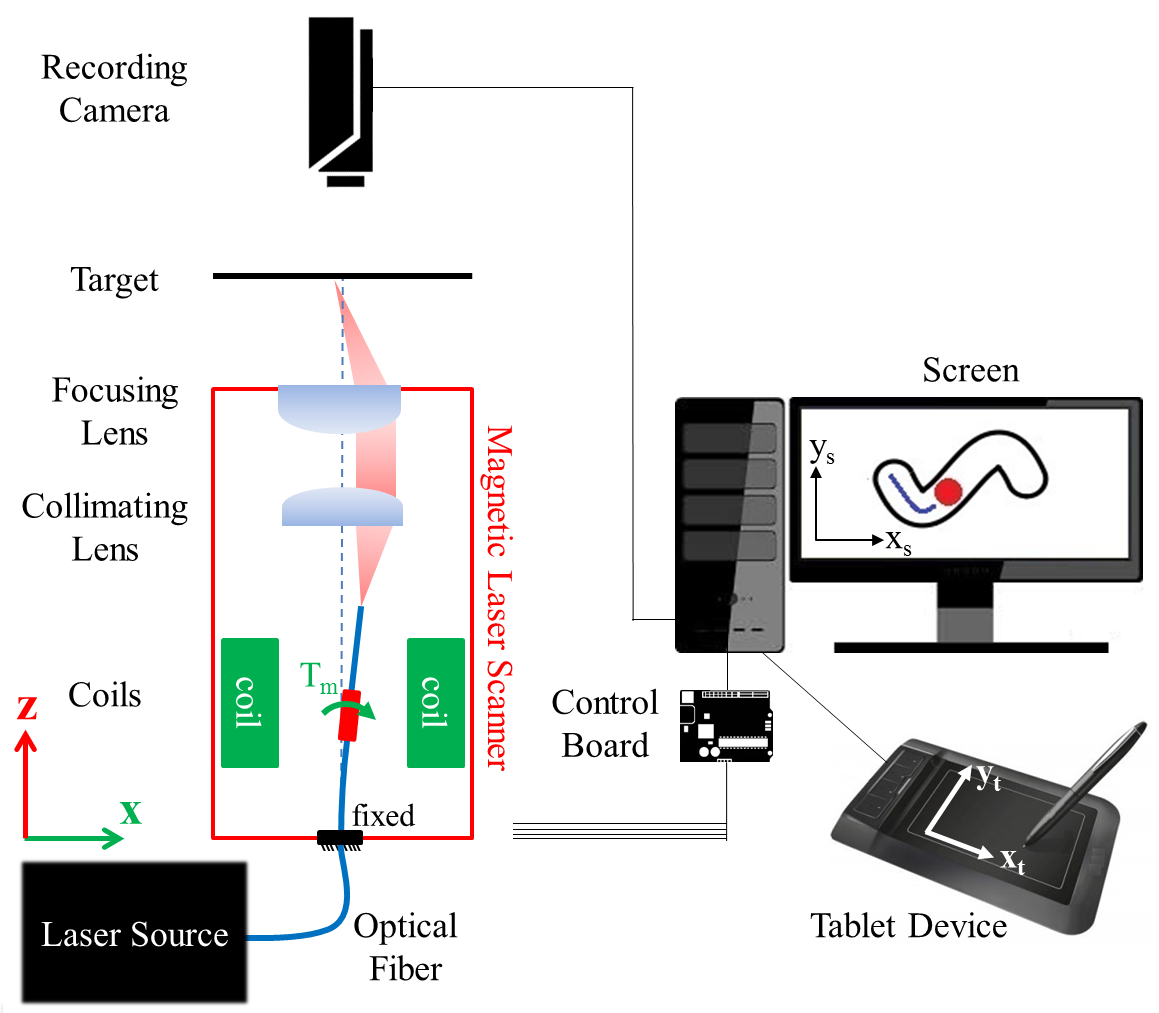}}
		\caption{Schematic illustration of the experimental setup. 
				 For simplicity, one degree of freedom is presented for magnetic actuation in the x-axis.
				 Laser spot position control on the computer screen with a tablet device is shown on the right side of the figure.}
			\label{fig_expset}
	\end{center}
\end{figurehere} 
The tablet is a mouse-like device that provides absolute position coordinates.
Its position values change between -1 and +1 in the x- and y- axes. 
The current control board provides 4096 levels for each current ($I_x$ and $I_y$).  
The mapping matrix scales the tablet positions [-1, 1] to the current control levels [-2047, 2048]. 
In the experimental setup, reference coordinate systems for magnetic laser scanner ($x$, $y$), screen ($x_s$, $y_s$) and tablet ($x_t$, $y_t$) are aligned to each other. 
This alignment implies that tablet x- position deflections affect only $I_x$ and tablet y- position deflections affect only $I_y$. 
In other words, coefficients $c_{12}$ and $c_{21}$ equal to 0.
$c_{11}$ and $c_{22}$ are equal, i.e., $c$.
In this experimental setup, $c$ value is defined as 2047.

\begin{equation}
	\label{eq_refraction_matrix1}
	\begin{bmatrix} 
            I_x\\
	        I_y    
	\end{bmatrix} 
        =  
    \begin{bmatrix} 
           c & 0\\
	       0 & c    
	\end{bmatrix} 
	\begin{bmatrix} 
             p_x\\
	         p_y     
	\end{bmatrix}.
\end{equation}

\subsection{Operating Modes}
The magnetic laser scanner scanner provides three operating modes, incorporating the features of state-of-the-art systems and offering an additional assistive feature for improved surgical experience:  

\begin{itemize}

\item \textit{High-speed scanning}: In this mode, the system actuates the laser to perform fast and continuous motions for improved quality of cuts. 
Characterization of high-speed scanning is performed with the \textit{deviation from linearity} experiments  described in \textit{Section      \ref{devlin}}.

\item \textit{Intra-operative incision planning}: During surgeries, surgeons generally define an incision path and execute this trajectory with repetitive manual passes. During these repetitions, keeping the laser spot exactly on the same path is highly important to avoid unexpected incisions in the vicinity of the desired path. The magnetic laser scanner allows surgeons to create incision paths intra-operatively and then the system can perform automatic repeated incisions on the same path. Characterization of this mode is presented with the \textit{repeatability} experiments  described in \textit{Section      \ref{repres}}.

\item \textit{2D position control}: In this mode, the laser spot position is controlled in the total workspace with free-hand motions on the tablet device. Characterization of this tablet control is done through \textit{teleoperation user trials} described in \textit{Section      \ref{teleop}}. 

\end{itemize}

\section{Experiments}

\subsection{Workspace}
In this experiment, the aim was to define the maximum limits of the workspace of the magnetic laser scanner by taking measurements from different locations.
To this end, a set of current values, $I_x$ and $I_y$, are fed to the coil pairs. 
Then, laser spot positions in the xy-plane and position($d$)-current($I$) correlations are calculated. 

\subsection{Deviation from linearity}
\label{devlin}
As mentioned earlier, the fast motion of the laser spot is crucial for clean incisions and minimal thermal damage to the surrounding healthy tissue.
The aim of this experiment is to measure the maximum errors observed during high-speed scanning. 
For this, the magnetic laser scanner is actuated with sinusoidal current inputs:
\begin{equation}
	\label{eq_refraction_matrix2}
		I = A \sin(2 \pi f t), 
\end{equation}
\noindent where $A$ is current amplitude, $f$ is excitation frequency and $t$ is time. 
The characterization experiments of the scanning feature was conducted by scanning a 0.72 mm straight line with different excitation frequencies up to 100 Hz. 
In order to observe the deviation from this linear path, a linear fit to the measurement points is accepted as reference.
Then RMSE values are calculated based on this reference for all excitation values.

\subsection{Repeatability}
\label{repres}
Repeatability is an important criteria for the laser microsurgery. 
During computer-controlled repetitions of predefined intra-operative trajectories,
the laser spot should travel exactly on the same path in order to guarantee accurate operation.   
This requirement becomes important when surgeons want to perform repeated ablations along the customized paths. 
In order to measure the repeatability of trajectories, a 2D trajectory with an  `8' figure is executed at 1 Hz and repeated 10 times. 
In this case, the first pass is accepted as the reference trajectory, allowing the computation of the RMSE for the subsequent passes of the laser spot.

\subsection{Teleoperation user trials}
\label{teleop}
The main objective of the proposed experiment is to analyse the usability and the 2D positioning accuracy of the magnetic laser scanner when controlled through a tablet device. 
To this end, user trials consisting trajectory-following tasks were performed.
12 subjects (11 male and 1 female, average age 27.25$\pm$3.5 years, 10 right-handed and 2 left-handed) performed 10 trials each with 5 different target trajectories resulting in 2 repetitions for each trajectory in a randomized order. 
They had no prior experience in 2D laser manipulation with a tablet device. 
All subjects received verbal and written information describing the experiments and its goals, 
then provided written informed consent in accordance with recommendations from
authors' institution and based on the Declaration of Helsinki.

Fig.~\ref{fig_shapes} depicts target trajectories (\textit{s-curves}, \textit{c-curves} and a straight line) used in the experiments.
In previous studies, laser position control systems were assessed by using similar trajectories with  the \textit{s-curves}, \textit{c-curves} and straight lines~\cite{deshpande2013imaging}.
These five different trajectories increase the variety of the shapes that can be used to assess the magnetic laser scanner.

In the experimental setup, the camera observes a white paper as target plane. 
The target trajectories are shown on the computer screen as virtual items created by changing the pixel colors of the video frames at the target locations~(Fig.~\ref{fig_expset2}).
Subjects are asked to follow the target trajectories keeping the laser spot inside of the borders of the shapes. 
The center position of the laser spot is compared to the central line of the target trajectory for assessing the control accuracy.

\vspace*{1cm}
\begin{figurehere}
	\begin{center}
		\centerline{\includegraphics[width=4.0in,trim={0 0 0 0}]{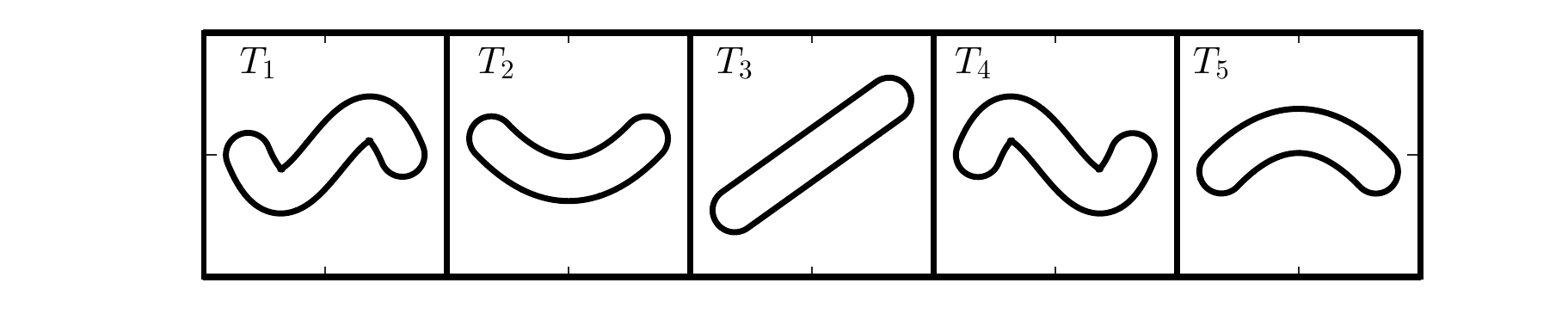}}
		\caption{The target trajectories that are used in the experiments: 
		$T_1$ and $T_4$ are \textit{s-curves},
		$T_2$ and $T_5$ are \textit{c-curves} and 
		$T_3$ is a straight trajectory.}
			\label{fig_shapes}
	\end{center}
\end{figurehere} 

\begin{figurehere}
	\begin{center}
		\centerline{\includegraphics[width=3in,trim={0cm 0.0 0 0}]{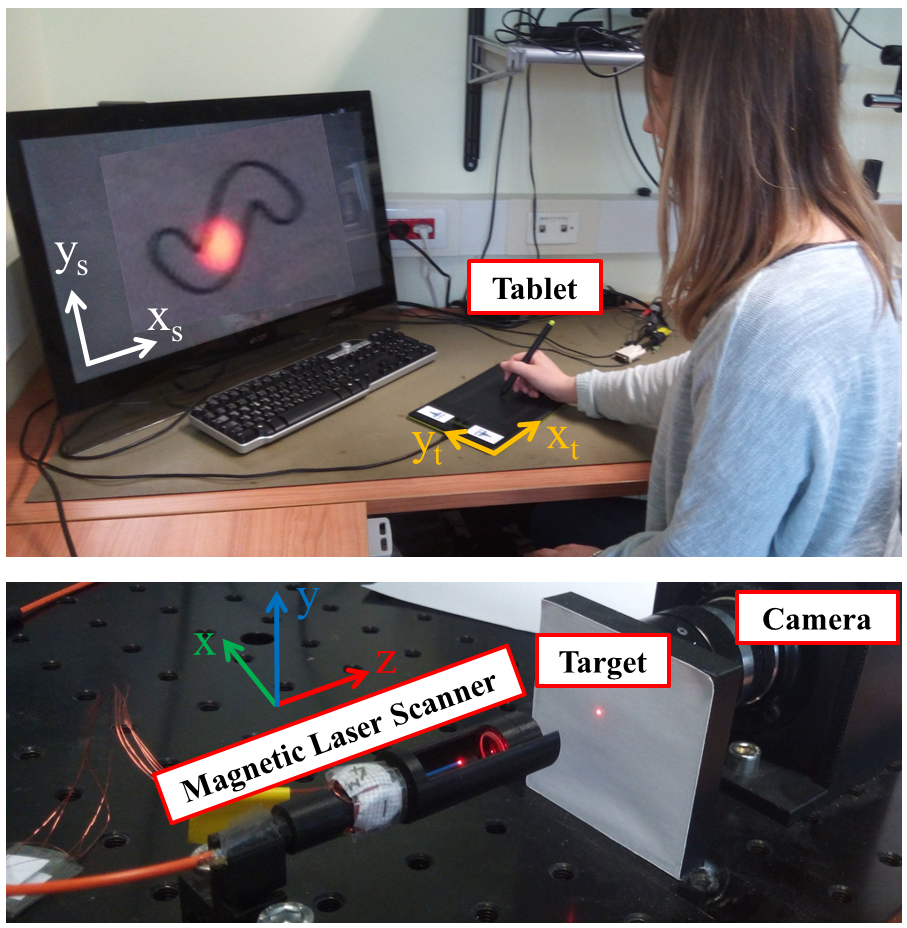}}
		\caption{A scene from user trials with the tablet control of magnetic laser scanner.}
			\label{fig_expset2}
	\end{center}
\end{figurehere}

\section{Evaluation Methodology}

In order to extract the laser positions for the deviation from linearity and repeatability experiments, videos are recorded with a high-speed camera (SpeedCam HiSpec 1 - Fastec Imaging) which is able to capture images at 1000 fps.
For the teleoperation user trials, the executed trajectories with a tablet device are recorded with a Prosilica GT1910C (Allied Vision) camera  at 25 fps as illustrated in Fig.~\ref{fig_expset}. 
The recorded color images are post-processed with MATLAB image processing tools in order to get the laser spot position in each frame.
The implemented laser detection algorithm processes the red channel of the image (since the laser used was red) and  is based on the intensity values of the pixels. 
Locations of pixel groups with intensity values above a certain threshold are recorded. 
Then, the centroid of the largest pixel group is accepted as the location of the laser spot. 
The threshold value is calibrated manually. 
It is defined between 0 and 1 depending on the illumination of the experimental setup. 

Pixel dimensions are measured using a piece of millimetric paper as target plane: 41 pixels = 1 mm. 
Thus, the measurements in pixel units are converted to mm with a scale factor of 24.39 $\mu$m/pixel. 
Separately, it is ensured that the entire region of the target trajectory is in focus and fully within the camera image plane.

For the workspace assessment, 
the initial laser position,
when no current is flowing through the electromagnet coils, is accepted as the reference point
([$I_x$, $I_y$] = [0, 0] $\longrightarrow$ [$d_x$, $d_y$] = [0, 0]).
This point is marked as a black circle in Figs.~\ref{fig_def_map} and \ref{fig_def_map1}.
Then, as the currents are changed, new laser positions are measured with respect to that reference point.

For the assessment of the deviation from linearity and repeatability experiments, the root-mean-square errors (RMSE) are calculated based on the reference trajectories. 
For the assessment of the teleoperation user trials, three metrics are used: 
		(i)   the RMSE based on the target trajectories,
		(ii)  the maximum errors, and
		(iii) the execution times.
The execution time is extracted from the recorded videos 
by measuring the time elapsed between the start of the execution of a trajectory until its completion.
For RMSE and maximum errors, the executed trajectories are compared with the target ones.
The errors are calculated by measuring the closest distance to the target trajectory.
For the $i^{th}$ measurement point, the error is:
\begin{equation}
   			err(i) = \min_{\forall k}(|d_m(i) - d_t(k)|),
			\label{eq_err1}
\end{equation}

\noindent where 
				$d_m$ is the executed trajectory and 
				$d_t$ is the target trajectory.
Maximum error is the highest value of $err$,
and root-mean-square-error (RMSE) is calculated as:
\begin{equation}
   \text{RMSE} = \sqrt{\frac{1}{N}\sum_{i=1}^{N}{{err(i)}^2}} ,
			\label{eq_rmse}
\end{equation}
\noindent where $N$ is the total number of measurement points.

\vspace*{1.5cm}
\textit{Subjective Analysis}: The aim of this analysis is to understand if the magnetic laser scanner can be effectively used to accomplish the specific tasks.
For subjective analysis of the teleoperation user trials, a dedicated questionnaire was prepared to assess the controllability and the response of the magnetic laser scanner.
The questionnaire, based on the System Usability Scale (SUS)~\cite{brooke1996sus}, focuses on 
				the ease-of-use of the system, 
				the fatigue induced by the control, 
				the mental work-load, 
				the accuracy and precision of the magnetic laser scanner, and 
				the intuitiveness of the control.
Table~\ref{tab_ques} presents the questionnaire items. 
In the questionnaire, some items have the same assessment objective, however they are expressed in a different way allowing to assess the consistency of the answers. 
At the end of the experimental session, subjects were presented this questionnaire on the computer screen and asked to fill out based on their experience with the device. 
Subjects chose their scores with a slide bar according to their agreement level with the questions (Disagree (1) - Agree (101)). 
Items 2, 4, 6, and 9 are expressed with a negative statement: disagreement is expected for these questions, whereas agreement is expected for the items 1, 3, 5, 7, 8, and 10. 
In addition to these items, an additional question was asked to the subjects: the perceived average execution time in seconds for a single trajectory.
This allowed the assessment of the perceived mental effort.

\begin{tablehere}
\tbl{Questionnaire Items \label{tab_ques}}
{\begin{tabular}[]{c | b{7.5cm}}
\toprule
1  & The control of the device (laser motion) was precise.\\ 
\hline
2  & The control of the device (laser motion) induced fatigue in my hand.\\
\hline
3  & The control/use of the device was easy.\\
\hline
4  & I was stressed, irritated or annoyed using this device.\\
\hline
5  & I found the device was easy to learn, so I could start using it quickly.\\
\hline
6  & The laser motion was not responsive enough to accomplish the task.\\
\hline
7  & The characteristics of the device (motion, speed, control) were sufficient to accomplish the task.\\ 
\hline
8  & My performance (accuracy) in this task with this device was satisfying.\\
\hline
9  & It is easy to make errors with this device.\\
\hline
10 & I would like to use this device again for this kind of task.\\
\hline
\end{tabular}}
\end{tablehere}

\section{Results}

\subsection{Workspace}
In order to establish the extent of the workspace, laser spot positions are measured for different sets of the current values, $I_x$ and $I_y$.
Fig.~\ref{fig_def_map} presents the 24 different laser spot positions for the current values varying between $\pm$0.165 A in x- and y-axes.
On the x- and y-axes, position of the laser spot is changing linearly with the current (Fig.~\ref{fig_def_map1}).
The current values result in a workspace almost 4$\times$4 mm$^2$ in size. 
\begin{figurehere}
	\begin{center}
 		\centerline{\includegraphics[width=3.5in,trim={0 0cm 0 0cm}]{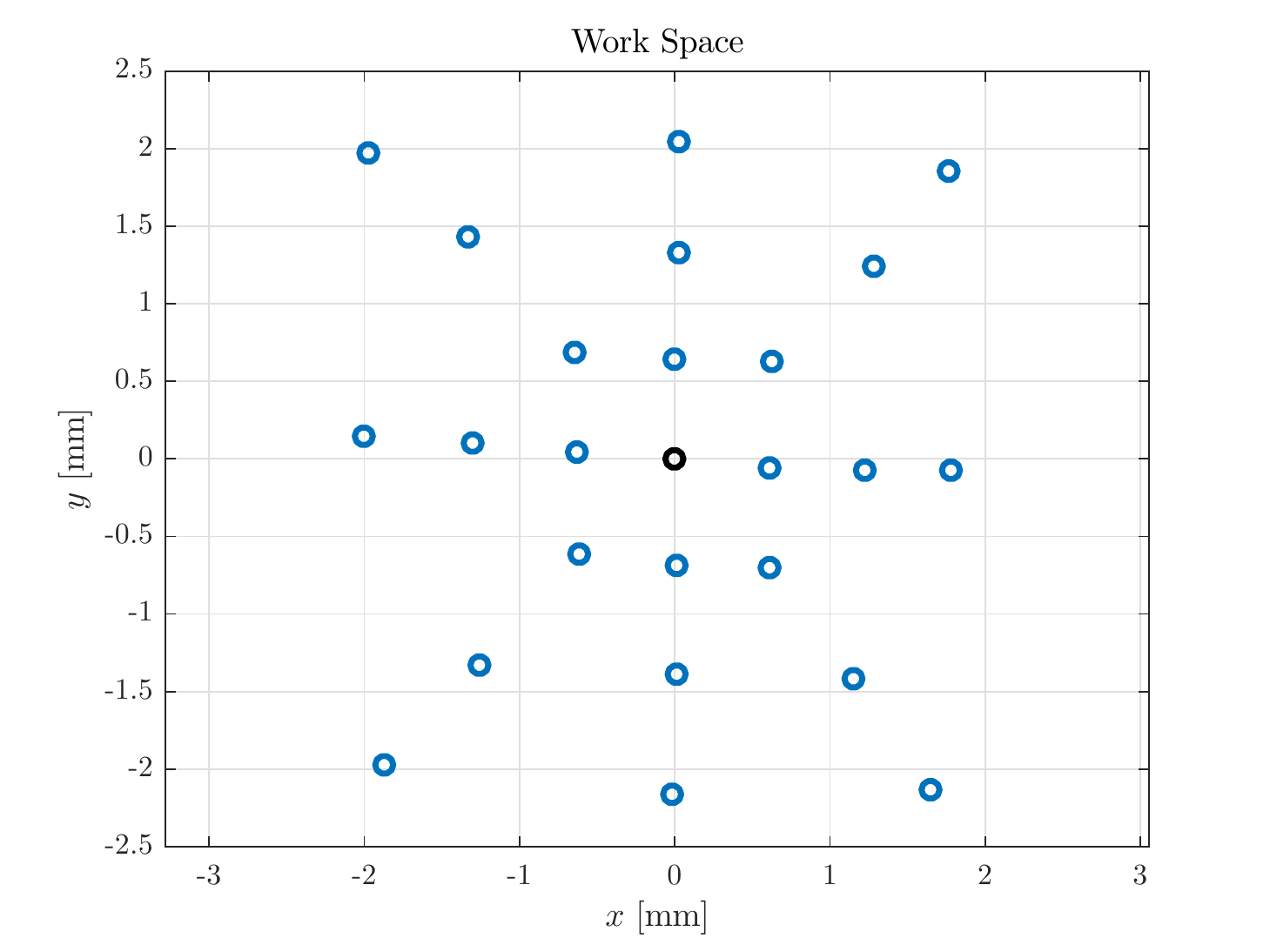}}
 		\caption{Laser spot positions for different current values on the target surface.		
 				\textit{Black circle} in [0, 0] represents the reference point, i.e., the initial laser position on the target for currents $I_x$ = 0 and $I_y$ = 0.
 				}
 				
 		\label{fig_def_map}
 	\end{center}
\end{figurehere}
\begin{figurehere}
	\begin{center}
 		\centerline{\includegraphics[width=3.5in,trim={0 0 0 0}]{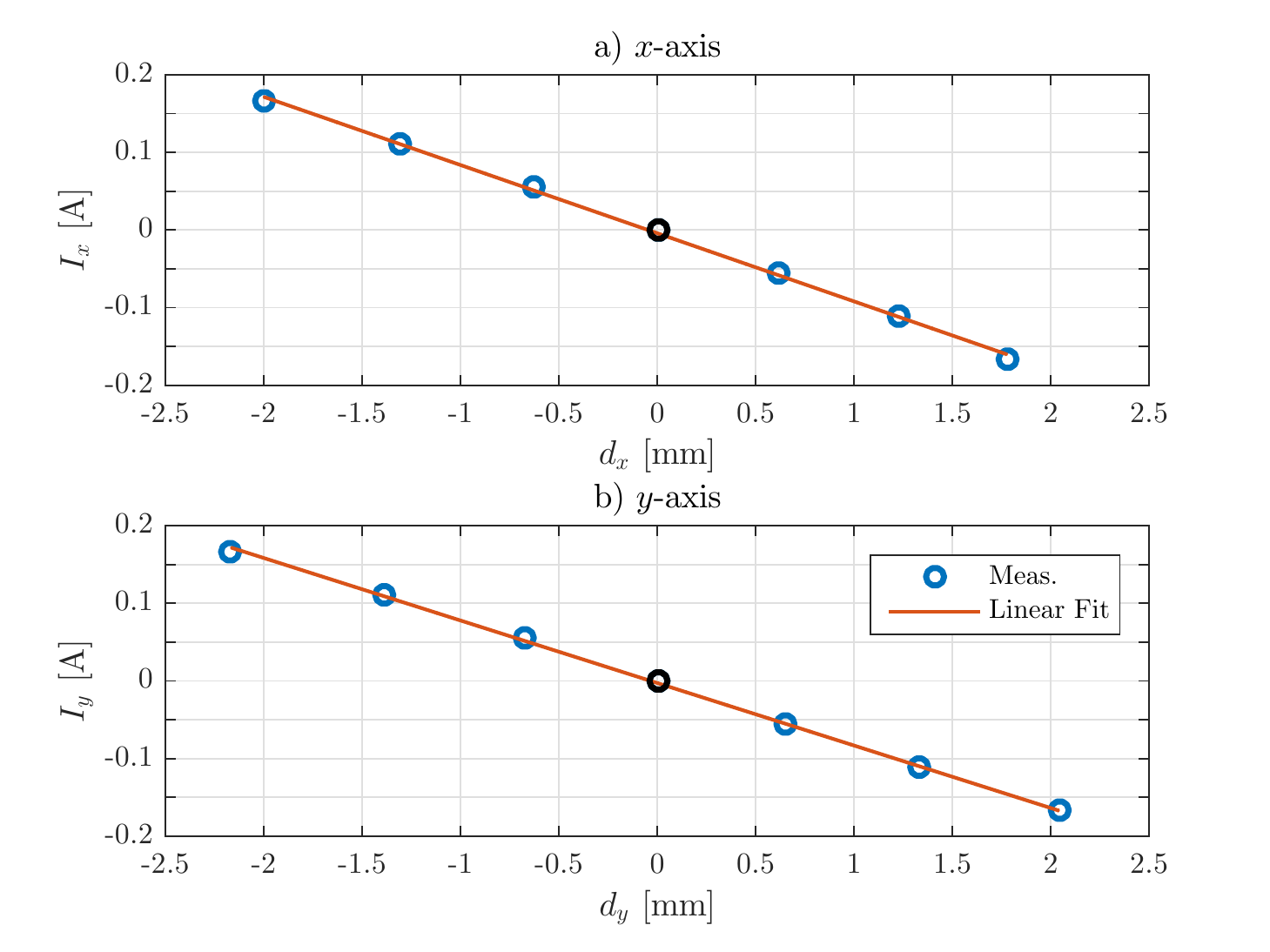}}
 		\caption{Current ($I$)- displacement ($d$) plots for x-axis (a) and y-axis (b).}
 		\label{fig_def_map1}
 	\end{center}
\end{figurehere}
In Fig.~\ref{fig_def_map}, it can be seen that laser spot positions are not perfectly aligned with each other in the workspace. 
These errors can be explained as resulting from manufacturing imperfections of the magnetic laser scanner. 
In the analysis, it is assumed that four identical electromagnetic coils are used.
However, independent impedance measurements have shown that there are 3-5\% differences among them.
Another assumption is that the coils facing each other on the same axis are placed perfectly parallel to each other. 
However, the mis-alignment of the coil pairs in x- and y-axes can cause different deflections on the positive and negative sides of the axes.  
These problems can be eliminated with better manufacturing and assembly of the device.

\subsection{Deviation from linearity}
In Fig.~\ref{fig_lindevab}a and b, a sample trajectory is shown for 5 Hz for the comparison of the measurement points to the linear fit.
For 5 Hz, RMSE value is 2 $\mu$m and maximum error is 7 $\mu$m. 
The same method is repeated for all excitation frequencies and 
RMSE values are presented in Fig.~\ref{fig_lindevc} with a linear frequency scale.
The natural frequency of the system is observed around 63 Hz. 
RMSE values increase exponentially as the excitation frequency is increased up to the natural frequency of the system.
Considering the thermal damage width values (about 50 $\mu$m) reported in previous research~\cite{remacle2008current}, 
the maximum stable scanning speed is chosen at 48 Hz for the errors less than 50 $\mu$m.
\begin{figurehere}
	\begin{center}
		\centerline{\includegraphics[width=3.5in,trim={0cm 0 0 0}]{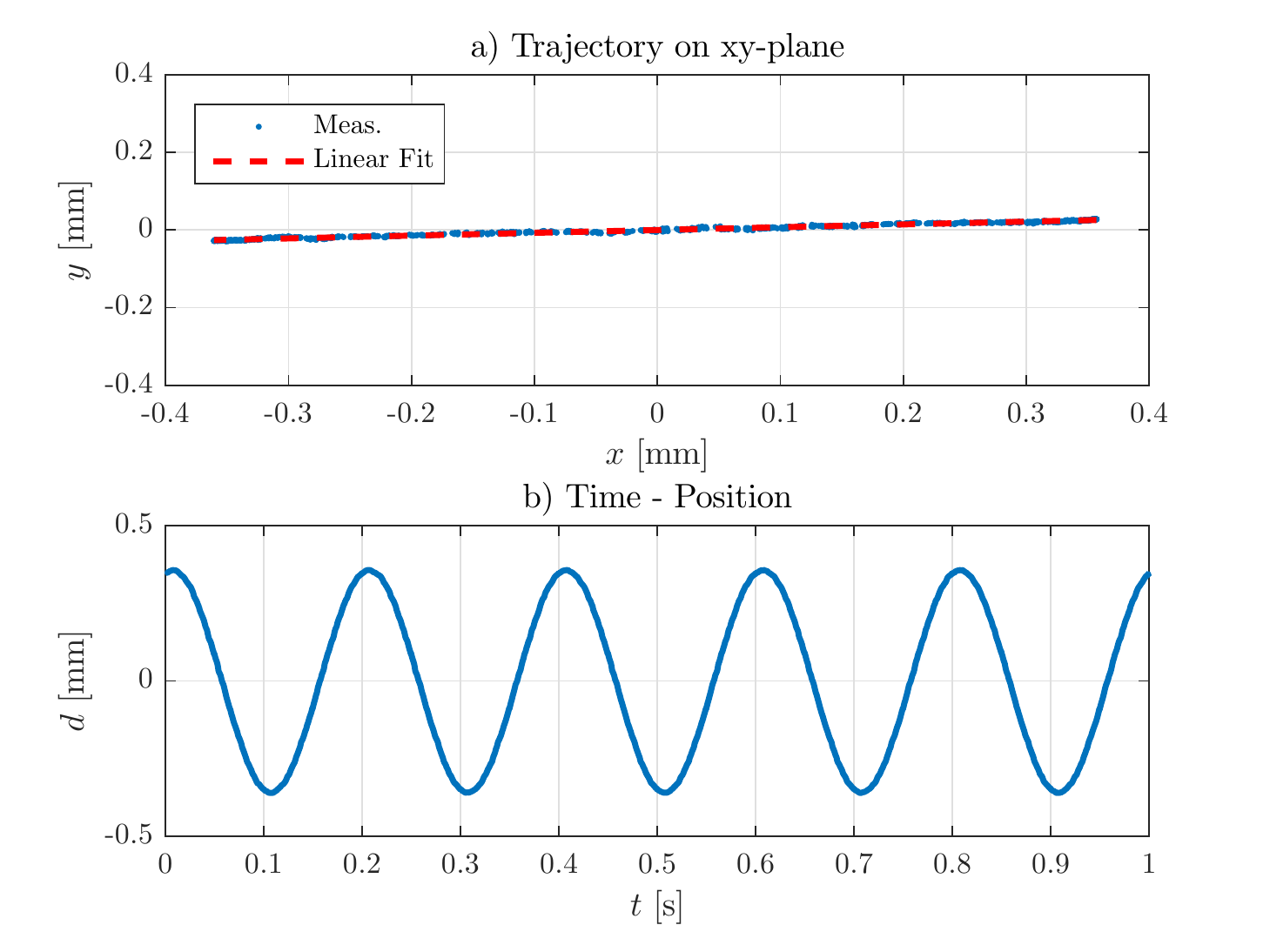}}
		\caption{a) A sample trajectory for RMSE calculation to demonstrate the deviation from linearity.
		         b) Time-position graph for the sample trajectory.
		 		}
			\label{fig_lindevab}
	\end{center}
\end{figurehere}

\begin{figurehere}
	\begin{center}
		\centerline{\includegraphics[width=3.5in,trim={0cm 0 0 0}]{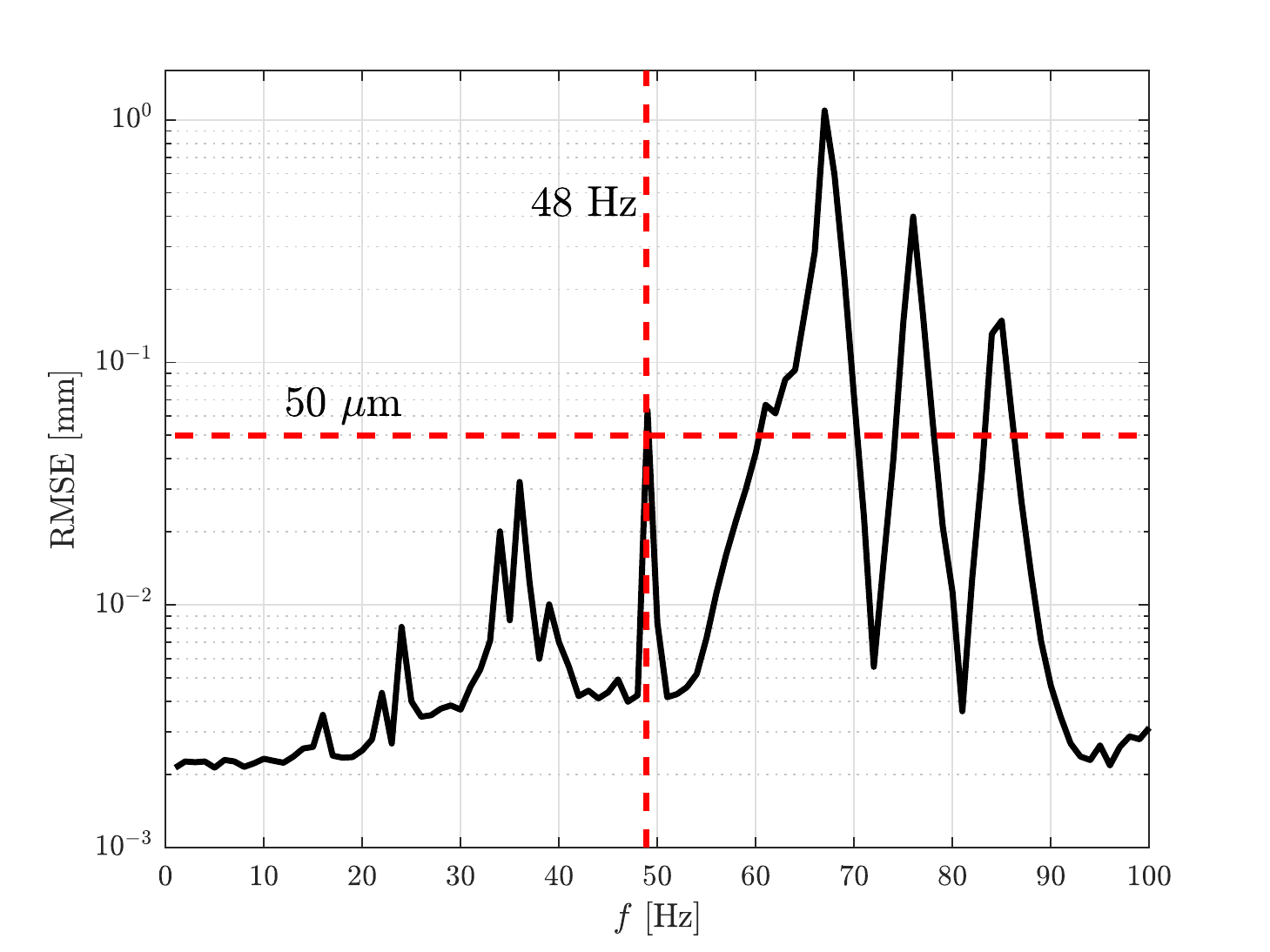}}
		\caption{Calculated RMSE for different excitation frequencies up to 100 Hz.
		         Average speed of the laser spot on the target was 1.44 and 144.2 mm/s for 1 and 48 Hz, respectively.}
			\label{fig_lindevc}
	\end{center}
\end{figurehere}

\subsection{Repeatability} 
\label{subsec_rep}

Fig.~\ref{fig_rep} depicts the results from the repeatability assessment considering the `8' figure (number of passes, $n$ = 10).
The calculated average RMSE for repeatability trials based on the first pass is  21 $\pm$ 10 $\mu$m.
Thus, the system shows very high repeatability  and is, therefore, suitable for precise intra-operative incisions during the surgery.

 \begin{figurehere}
	\begin{center}
		\centerline{\includegraphics[width=3.5in,trim={0cm 0 0 0}]{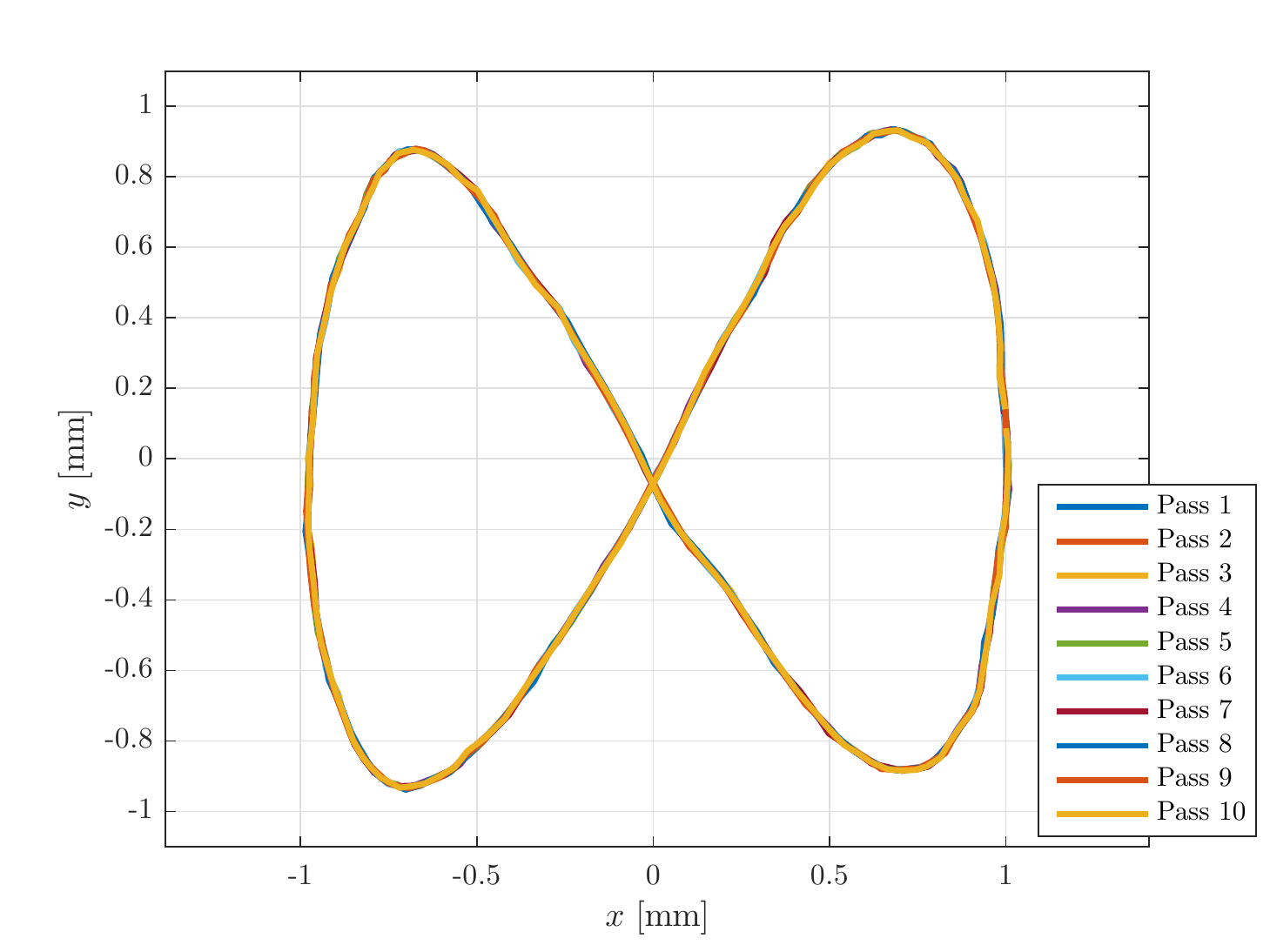}}
		\caption{Repeatability results of the scan trajectories for 10 passes. 
				All trajectories are overlapping and indistinguishable from each other.			
				} 
			\label{fig_rep}
	\end{center}
\end{figurehere}
\begin{figurehere}
	\begin{center}
		\centerline{\includegraphics[width=3.5in,trim={0cm 0 0 1cm}]{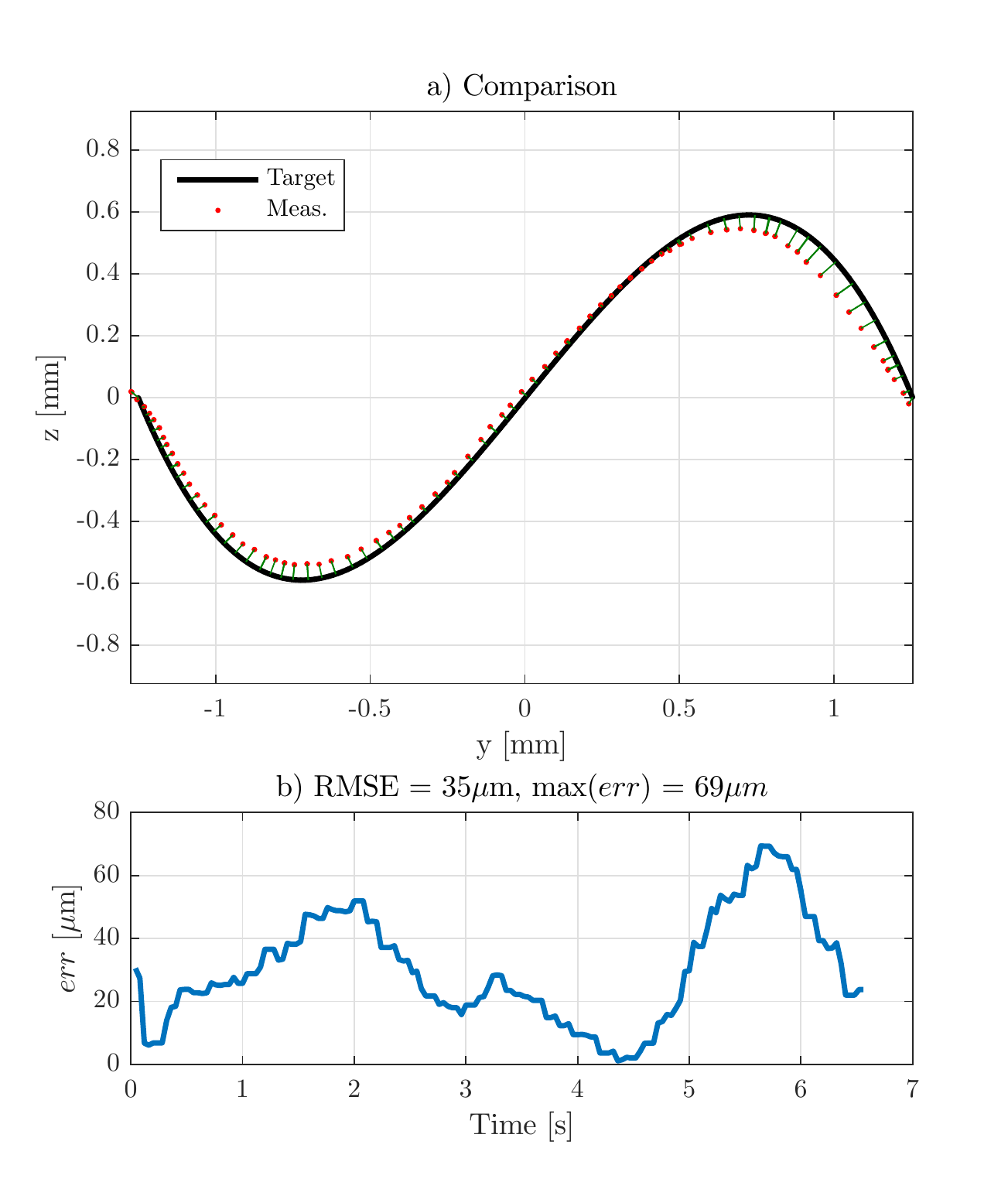}}
		\caption{a) Comparison of a sample \textit{s-curve} trajectory (\textit{red dots}) against a target trajectory (\textit{black line}).  
					\textit{Green lines}  show the error bars for each measurement point.
				 b) Calculated errors for all measurement points for the sample trajectory.}
			\label{fig_rmsecalc}
	\end{center}
\end{figurehere}
\subsection{Teleoperation user trials}
Fig.~\ref{fig_rmsecalc}a shows a comparison of the executed and target trajectories with the error bars.
Fig.~\ref{fig_rmsecalc}b presents the errors for this trajectory and calculated RMSE. 
Fig.~\ref{fig_subrmse} depicts the RMSE for each subject over 10 trials. 
The average RMSE over the 12 subjects is 39 $\pm$ 8 $\mu$m.
These values indicate that the magnetic laser scanner also provides good accuracy when compared to the traditional systems with the manual micromanipulator considering the previously reported RMSE values (211 - 650 $\mu$m) in the literature~\cite{mattos2011virtual, mattos2014novel, deshpande2014enhanced}.

\begin{figurehere}
	\begin{center}
		\centerline{\includegraphics[width=3.5in,trim={0cm 0 0 0}]{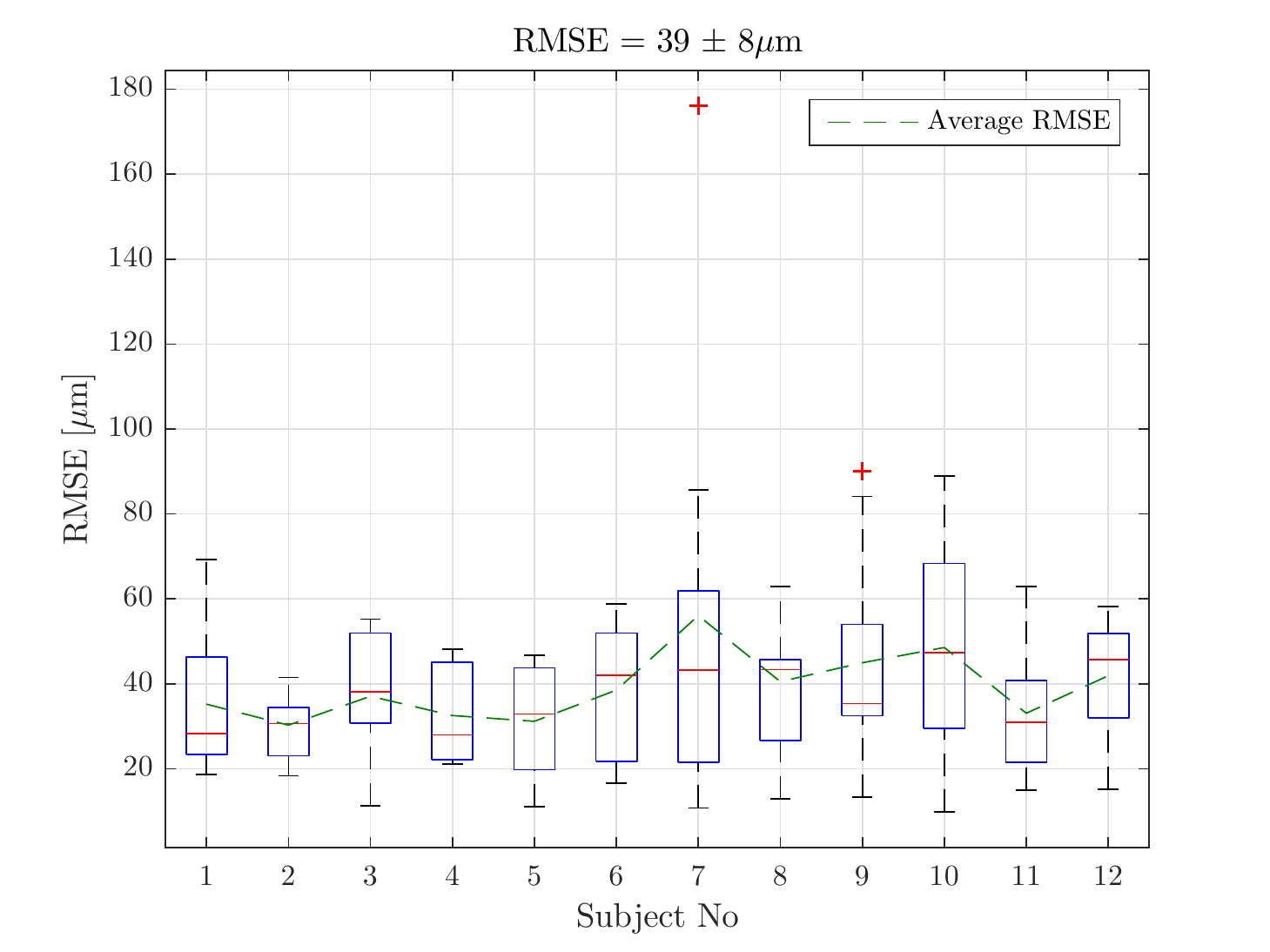}}
		\caption{The boxplots of the RMSE for each subject over 10 trials. The outliers are plotted individually with the \textit{red} `+' symbol.}
			\label{fig_subrmse}
	\end{center}
\end{figurehere}
\begin{figurehere}
	\begin{center}
		\centerline{\includegraphics[width=3.5in,trim={0cm 0 0 0}]{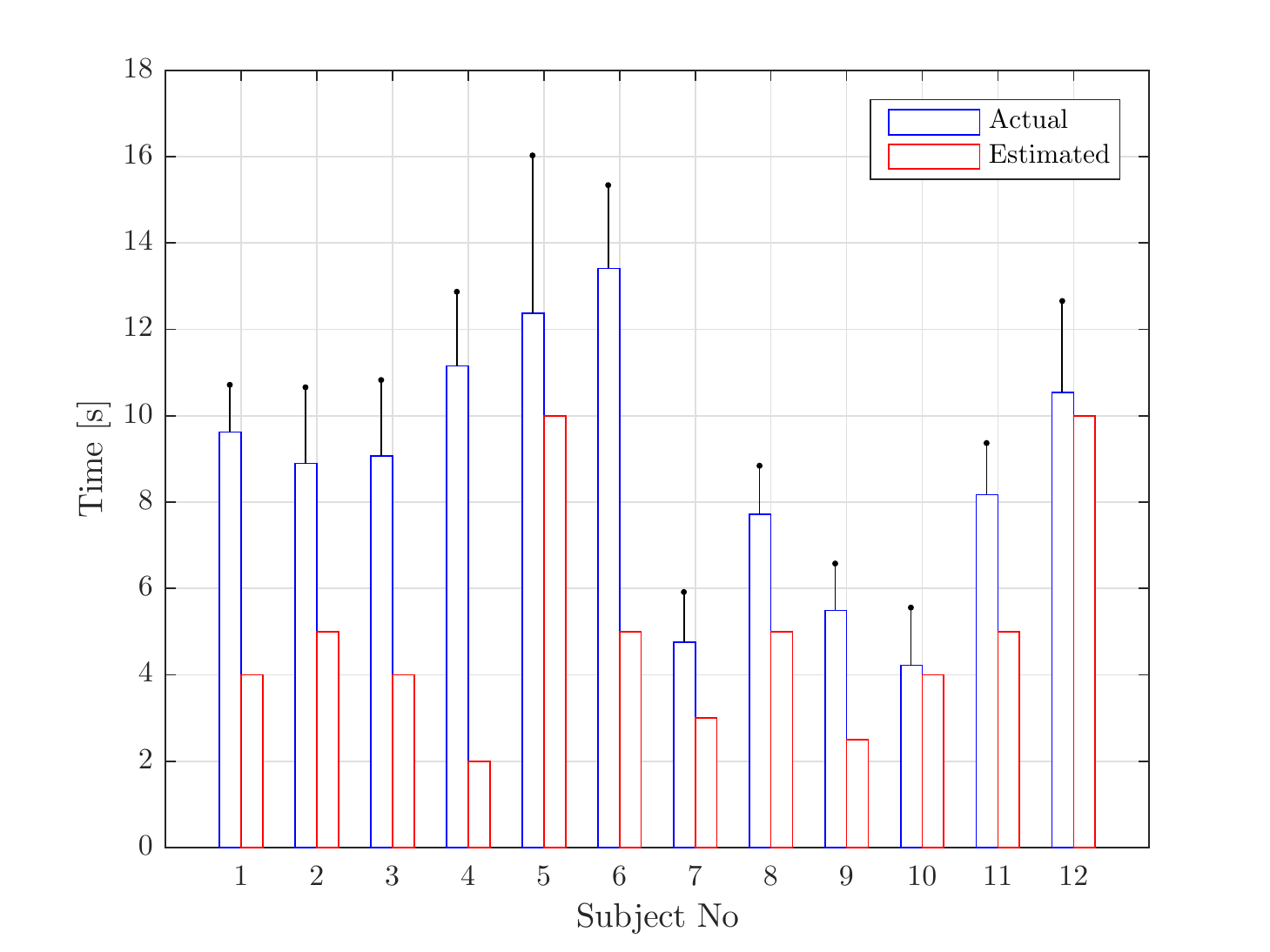}}
		\caption{Comparison of the actual (\textit{blue}) and estimated (\textit{red}) average execution times.}
			\label{fig_subtime1}
	\end{center}
\end{figurehere}
Fig.~\ref{fig_subtime1} depicts the time required to follow the target trajectory for each subject.
The average time for all subjects is 8.78 seconds. 
Taking into account the RMSE values (Fig.~\ref{fig_subrmse}) and trajectory execution times (Fig.~\ref{fig_subtime1}),
it can be observed that slower trajectory executions lead to lower errors. 
For instance, subjects 2 and 4 executed the trajectories with average times of 8.9 and 11.2 seconds, respectively. 
The corresponding RMSE values are 30 and 33 $\mu$m.
This shows that the system allows precise control of the laser spot. 
Thus, surgeons can carefully execute incisions paths precisely on the desired locations.
Moreover, fast trajectory executions tended to lead to larger errors as compared to the slow ones. 
The largest errors were observed for subjects 7, 9 and 10, who presented also the lower average trajectory execution times
(RMSE values equal to 56, 45 and 48 $\mu$m and times equal to 4.8, 5.5 and 4.2 seconds, respectively).
In order to assess the learning during the experiments, Figs.~\ref{fig_trialrmsemax} and~\ref{fig_trialtime} are presented for error and time evaluation for each trial. 
The average values of the RMSE and maximum errors are calculated over 12 subjects for each trial and results are presented in Fig.~\ref{fig_trialrmsemax}.
RMSE and maximum errors have similar behavior and both decrease with the trial number. 
The maximum error for the first trial is 108 $\mu$m and for the last trial, 71 $\mu$m.
For the average RMSE, first and last values are 49 and 33 $\mu$m, respectively. 
The general behavior of the error shows that subjects are improving their precision during the experiments. 
Fig.~\ref{fig_trialtime} depicts the average time evaluation over 10 trials. 
The results show that the average time decreases by 1.1 seconds. 
The system allows increased ease-of-use even with a little training.
\begin{figurehere}
	\begin{center}
		\centerline{\includegraphics[width=3.5in,trim={0cm 0 0 0}]{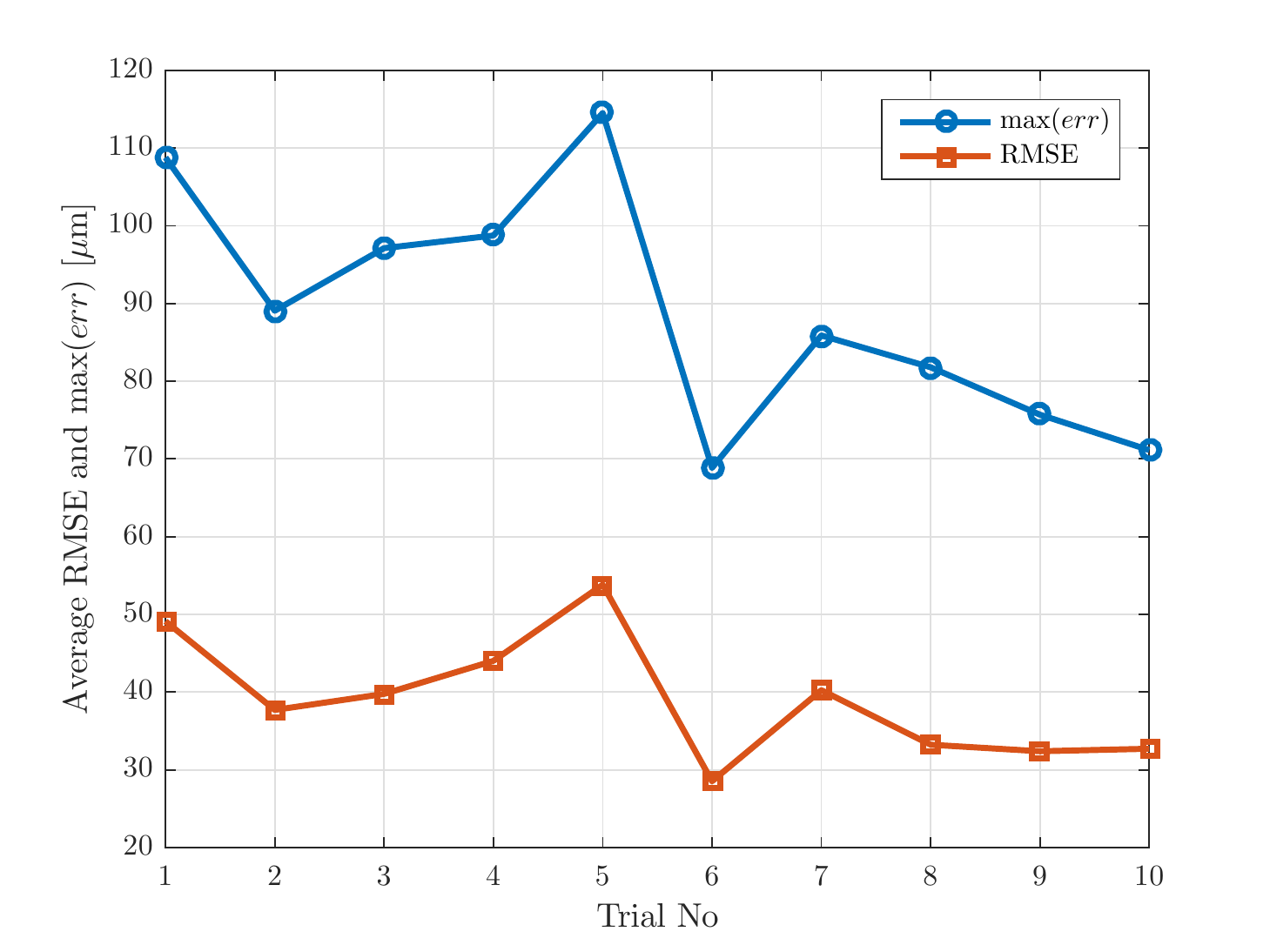}}
		\caption{Overall average RMSE and maximum errors for each trial considering the data from all experimental subjects.}
			\label{fig_trialrmsemax}
	\end{center}
\end{figurehere}
\begin{figurehere}
	\begin{center}
		\centerline{\includegraphics[width=3.5in,trim={0cm 0 0 0}]{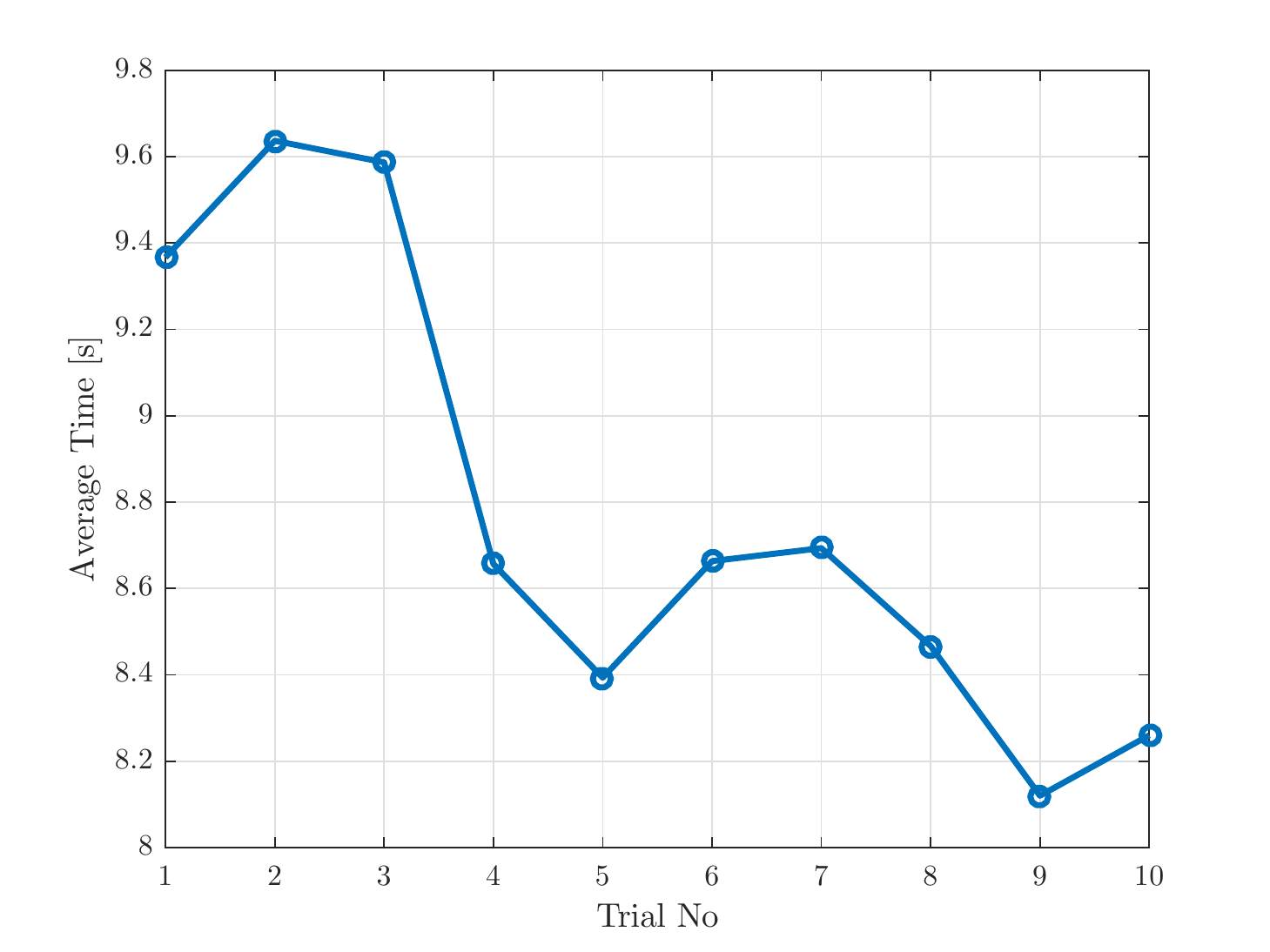}}
		\caption{Average time for each trial over 12 subjects. }
			\label{fig_trialtime}
	\end{center}
\end{figurehere}

\textit{Subjective Analysis}: Fig.~\ref{fig_quesres} depicts the questionnaire results for the items presented in Table~\ref{tab_ques}.
It is observed that the overall assessment is very positive for the device.
Specifically, the following points are noted:
\begin{itemize}
\item \textit{The ease-of-use}: High scores on items 3, 5, and 10 indicate that the system is easy to use and can be learned easily.

\item \textit{The fatigue}: Low scores on items 2 and 4 shows that the control method does not induce fatigue in the hand.

\item \textit{The mental work load}: The scores of the items 3, 4, and 5 indicate that the control of the system does not require high mental effort. The required cognitive work load is not high.

\item \textit{The accuracy and precision}: The items 1 and 7 have high scores  and the item  6 and 9 low scores: indicating that users assess the magnetic actuation as accurate and precise for the execution of laser positioning tasks.

\item \textit{Intuitiveness}: The scores of the items 3, 4, and 10 indicate that the control of the magnetic laser scanner is intuitive.

\item \textit{Self-assessment}: The high score of the item 8 shows that subjects felt confident in using the system.
\end{itemize}

The estimation of elapsed time to complete a specific task can provide an indication of the perceived mental workload for that task~\cite{lind2007time}. 
For self-assessment of this perceived mental workload, subjects were asked to estimate an average execution time for their performances after 10 trials. 
In Fig.~\ref{fig_subtime1}, actual execution times are compared with the estimated execution times. 
As can be seen, the estimated execution times are lower than the actual execution times.
This indicates that, even though the actual effort may be high, the perceived mental effort to use the system is low.
This is a positive assessment for the system.

\begin{figurehere}
	\begin{center}
		\centerline{\includegraphics[width=3.5in,trim={0cm 0 0 0.2cm}]{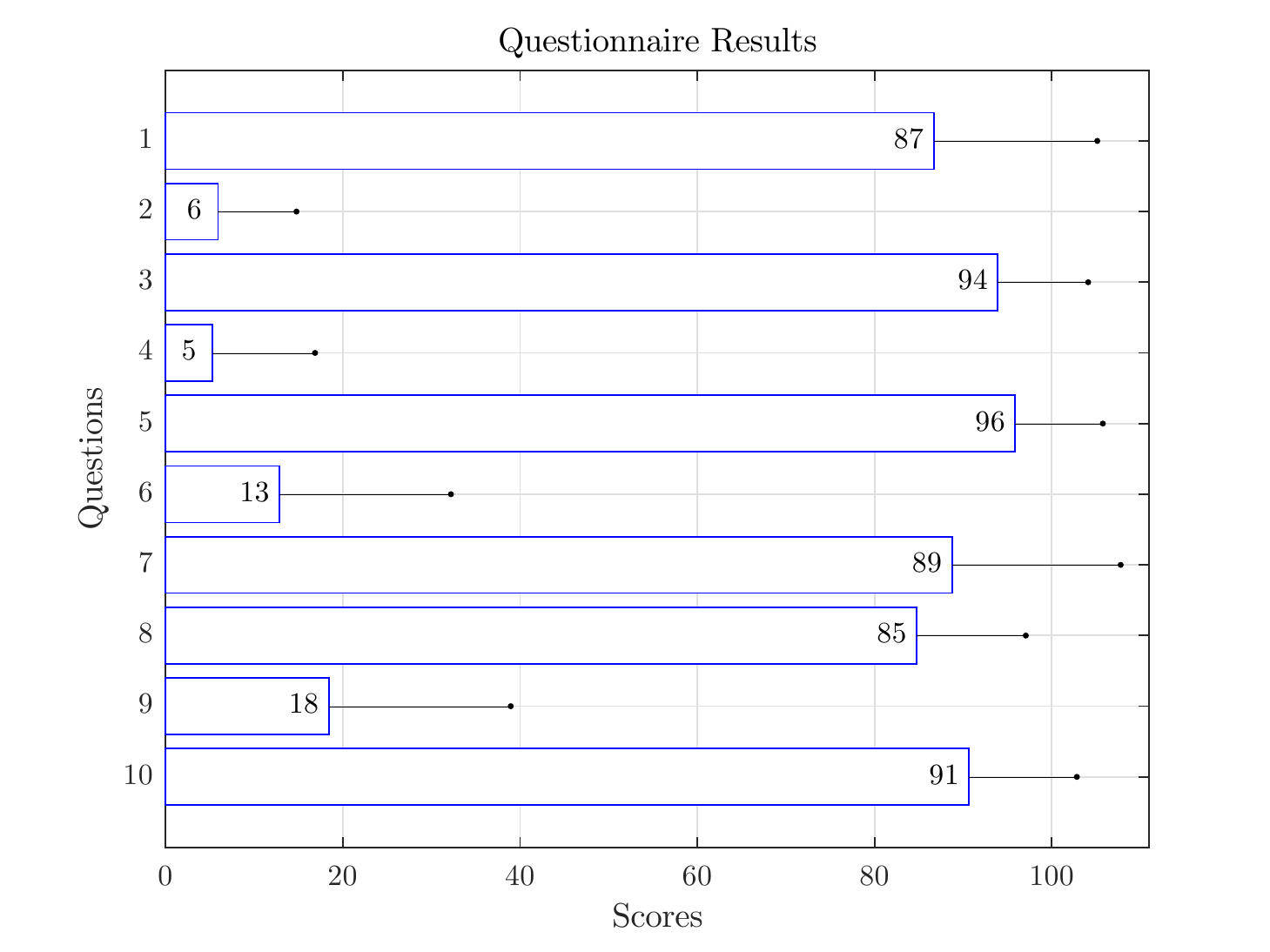}}
		\caption{Questionnaire results}
			\label{fig_quesres}
	\end{center}
\end{figurehere}

\section{Discussion}

The results presented above show that the new concept of teleoperated endoscopic laser manipulation improves the accuracy of the laser control compared to the state-of-the-art laser manipulation technique based on manual micromanipulators.
Accuracy assessment of operations with the magnetic laser scanner shows that average RMSE on trajectory tracing tasks are less than 40 $\mu$m.
This value compares positively with respect to RMSE values in the range 211 - 650 $\mu$m reported for the manual micromanipulator with 400 mm operating distance~\cite{mattos2011virtual, mattos2014novel, deshpande2014enhanced}.
Thus, endoscopic laser control provides at least five times better accuracy than traditional laser manipulation techniques due to: 
1) shorter operating distance; 
2) high-resolution laser micromanipulation capabilities; and 
3) the use of teleoperation techniques, including motion scaling and better user interface.

In addition, when using the magnetic laser scanner, surgeons can define a customized trajectory with the tablet device, which is then automatically executed by the system with high accuracy and precision, allowing repeated ablations along the same path.
Results from precision assessment show that repetitions of a predefined trajectory are indistinguishable from each other.
Additionally, the repeatability error is around 20 $\mu$m which is within the range of thermal damage caused by CO$_2$ lasers (about 50 $\mu$m)~\cite{remacle2008current}.

The total workspace of the prototype magnetic laser scanner is 4x4 mm$^2$.
Prior research indicates that surgeons prefer the high speed scanning lengths  in the range of 1-2 mm~\cite{remacle2005new, remacle2008current, fiorelli2013digital}.
However, commercial systems with mirror-based scanning provide incision lengths up to 5 mm~\cite{remacle2008current}.
Therefore, the achieved total workspace is comparable to the state-of-the-art systems. 
The workspace of the magnetic laser scanner can be increased further by adapting the optical design for longer working distances between the target and the tip of the scanner. 
The extent of the workspace would increase linearly with the increasing focal length. 
Additionally, coupling the magnetic laser scanner to the distal end of a flexible robotic endoscope would also increase the workspace by enabling the motion of the end-effector module itself. 
However, when adapting the optical design and integrating the system with a flexible endoscope, the total volume of the surgical site should be considered as a design restriction.

\section{Conclusion}

In this paper, the design and control of a novel magnetic laser scanner is presented to be used in endoscopic microsurgery.
The system is designed as a tip module of an endoscopic system in order to provide 2D position control and high-speed scanning of a surgical laser.
The prototype involves the actuation of an optical fiber using the magnetic interaction between electromagnetic coils and a permanent magnet.  
Performance of the proof-of-concept device has been assessed with characterization experiments and teleoperation user trials.
The magnetic laser scanner demonstrated to be able to perform stable scanning motions up to 48 Hz, and to provide high repeatability with an error around 20 $\mu$m.
In addition, the prototype has been integrated with a tablet device for teleoperation.
The accuracy of such integrated system has been assessed through user trials involving 12 subjects.
Results showed that target trajectories can be executed with 39$\pm$8 $\mu$m accuracy, which is at least five times better than reported values regarding current state-of-the-art laser systems.
Furthermore, subjective assessments indicated that the proposed system is intuitive and easy to control, presenting good response to tablet commands. 
In ongoing efforts, the system will be coupled to a surgical laser for the characterization of the tissue ablations leading to an important validation towards real clinical use of the proposed technology.\\

\bibliographystyle{ws-jmrr}
\bibliography{main}

\end{multicols}

\end{document}